# DECISION TREE INDUCTION SYSTEMS: A BAYESIAN ANALYSIS

Wray Buntine
(sesg.oz!wray@seismo.css.gov)

*Affiliation*: N.S.W. Institute of Technology and Macquarie University
Centre for Advanced Computing Sciences
New South Wales Institute of Technology
PO Box 123, Sydney, 2007, Australia

**ABSTRACT**

Decision tree induction systems are being used for knowledge acquisition. Yet they have been developed without proper regard for the subjective Bayesian theory of inductive inference. This paper examines the problem tackled by these systems from the Bayesian view in order to interpret the systems and the heuristic methods they use. It is shown that decision tree systems depart from the usual Bayesian methods by implicitly incorporating prior belief that the simpler of two hypotheses will be preferred, all else being equal. They perform a greedy search of the space of rules to find one in which there is strong posterior belief.

## 1. INTRODUCTION

A common task in knowledge-based systems is inducing a rule to handle simple binary classification. Given a set of classified (positive and negative) examples of some concept, the task is to develop a classification rule to predict the class of further unclassified examples. It is performed by induction systems as an aid to knowledge acquisition [Michalski 1983, Quinlan, Compton, Horn and Lazarus 1986] when examples are available but more general knowledge is not.

Decision tree induction systems are particularly good at this task. There are two independently developed families of systems, first, ID3 [Quinlan 1986a], ACLS [Paterson and Niblett 1983] and its commercial derivatives, and secondly, the CART suite of programs [Breiman, Friedman, Olshen and Stone 1984]. Both families have a string of industrial and academic successes to their credit [Breiman *et al.* 1984, Quinlan 1986a, 1986b].

These systems can be viewed from two key vantage-points. The first is from the cognitive scientist's view. The following questions are indicative. What is the cognitive environment in which the system operates? How should the system best facilitate knowledge acquisition from an expert? The second is from the statistician's view: what is the formulation of the problem and how should the system make the best decision? Both aspects are vital to the understanding and engineering of a successful system.

This paper is cast in the statistician's view. No rational argument can deny that the subjective Bayesian theory of induction provides a correct framework for the task these systems are tackling [Horvitz, Heckerman and Langlotz 1986]. Not one of these systems, however, is based on the Bayesian theory.

I believe it is important to resolve this apparent conflict between theory and practice in order to clarify and further develop these pragmatic systems. This may allow us, for instance, to consider how they may be improved or what their current scope of application is. Also, I believe it is important to elucidate the role that statistics plays in understanding operational induction systems, so that other less precise issues (such as the cognitive environment in which they operate) can also be understood.

The exposition proceeds as follows. After discussing the induction task in more detail, I review the Bayesian approach, assuming examples come from a (possibly culled) historical database. Within this framework, I investigate implications of the knowledge acquisition context to this approach, in particular, what form of prior may be suitable. I then discuss and interpret the decision tree induction systems ID3 and ID3 with pruning [Quinlan 1986b, Quinlan *et al.* 1986].

## 2. THE INDUCTION TASK

For the induction task in the knowledge acquisition context, we typically have an *expert* who is sufficiently knowledgeable to formulate the problem for us and is in possession of a *training set* (a set of objects whose classification are known). The task is to develop a *classification rule* to predict the *class* of further, unclassified examples. I consider here only two-class induction: each object is either a *positive* or *negative* instance of some concept under study.

The problem formulation is done as follows. The *objects* are grouped into different *types*. In a given problem a particular type of object is usually associated with a particular *description* in terms of an expert-supplied *language*, consisting of, say, 10-30 *attributes*. Each attribute may be binary ("true" or "false"), multi-valued, or real-valued [Quinlan 1986b]. Only the simple binary case is considered here.

Quinlan *et al.* [1986] present an induction problem where objects correspond to patients that attended a laboratory for endocrine analysis. Each patient is described in terms of attributes such as sex, age, pregnant and on-lithium. Two patients are considered to be of the same type if they have the same attribute values. One binary classification of patients is whether they are hypothyroid or not hypothyroid.

The training set available for this problem is a set of some 4000 recent medical records. Older records are not comparable as different or less accurate measurements were recorded. In the knowledge acquisition context training sets rarely have more than several thousand examples, several hundred is more common. This is because the economics or time-scale of a problem usually limits the amount of data available.

A critical part of the decision tree approach to induction is the selection of the language (the attributes) [Michie 1986a]. The attributes chosen are often ones that experts consider useful for classification. For instance, they could be key primitives on which they traditionally base their decisions, or attributes that are relevant according to some causal or incomplete theory. Michie claims that in practice experts can often articulate useful attributes when they do not have a sufficient mental grip on their task to articulate rules.

A *classification rule* can be selected on a number of grounds. In the above hypothyroid example we may be required to decide whether to treat a patient, or to predict the odds against a patient being hypothyroid. The "best" rule may be one that performs its function with a high *predictive accuracy*. But with knowledge-based systems where accountability is important and experts seek to understand or assess rules themselves, rules also have to be *comprehensible* [Michie 1986a]. This is a cognitive constraint. A huge lookup table of decisions indexed by thousands of different types of objects could not be comprehensible due to its complexity.

A knowledge engineer would typically guide the expert in a cyclic process of performing induction, evaluating the worth of rules, and perhaps modifying the problem formulation or augmenting the training set.

## 3. THE STATISTICIAN'S VIEW

Rule induction is in fact a generalization of the simple Bernoulli trial, treated in most elementary statistical test books. A common example is the public opinion poll. Suppose you want to decide whether a majority of people support trade-based unions and have posed the question to a random sample of 150 people. How do you make an educated guess given this data? This is a binary classification problem where, in the terminology above, there is only a single type of object, the "person", and the classes are "support" and "do not support".

The subjective Bayesian approach to this simple problem has been analysed in depth by Howard [1970]. Induction is decomposed into two processes, belief analysis to determine *posterior belief* followed by decision analysis to obtain the *best decision* [Raiffa and Schlaifer 1961].

### Assumptions

To determine posterior belief, the way the training set is obtained and prior belief need to be modelled. I shall consider the situation where the training set represents a historical database, perhaps modified to include sufficient examples of rare cases and to exclude excessive redundancy of common cases. The following assumptions apply:

(1) the distribution of objects and their class does not vary over time,
(2) objects in the training set are drawn from the population independently of their class,
(3) the training set is completed in a non-informative manner [Raiffa and Schlaifer 1961], and
(4) prior belief about class given type is independent of prior belief about the distribution of object types.

For instance, objects could be selected to ensure the training set includes a representative coverage of the different possible types and assumption (2) still holds as long as the selection method is not influenced by the classification of the objects. Assumption (3) would not apply if the expert considered the training set sufficient to demonstrate the concept under study, or had hand-crafted the training set as a tutorial set of examples [Michie 1986a, 1986b]. Prior belief is further discussed in the next section.

To model the decision-making process assume the following scenario. In the available training set class corresponds to an expert's (or the best-in-hindsight) yes/no decision about objects [Michie 1986b]. The system is required to emulate this example decision-making as closely as possible, and provide a concise rule explaining its behaviour. The system could attempt to minimize the expected number of errors in classification. This, of course, is only suitable when the payoff structure (the relative cost of success and failure) is reasonably constant. Furthermore, the concern for comprehensibility has been ignored. Nevertheless, I shall adopt this criterion here as it happens to be a common measure on which decision trees are judged.

### Notation

Let $C$ be the number of different types of objects. For instance, with 15 binary-valued attributes there are $2^{15}$ different descriptions we can assign to an object. That is, $C \approx 32,000$. For $i=1..C$, let $\lambda_i$ represent the proportion of objects of type $i$ that would occur in a very large number of examples and $\phi_i$ represent the proportion of those objects of type $i$ that are positive instances. These are well defined by assumption (1). So $(\lambda_i \phi_i)$ represents the proportion of positive instances of objects of type $i$, $(\lambda_i(1-\phi_i))$ represents the proportion of negative instances, and $\sum_{i=1..C} \lambda_i = 1$. Let $p_i$ be the number of positive instances of the $i$-th type in the training set and $n_i$ be likewise for negative instances. Let the size of the training set be equal to $n$, so $\sum_{i=1..C}(p_i+n_i)=n$.

### Analysing Belief

Current belief about objects and their classification can be represented by a density function on $(\underline{\lambda},\underline{\phi})$, vectors of the real numbers $\lambda_i$ and $\phi_i$. $\underline{\phi}$ would parameterize a particular classification rule and $\underline{\lambda}$ would parameterize how different types of objects are distributed. By Bayes theorem and assumptions (2), (3) and (4), for the marginal posterior belief about $\underline{\phi}$

$$Posterior(\underline{\phi}) \propto Prior(\underline{\phi}) \cdot \prod_{i=1..C} \phi_i^{p_i}(1-\phi_i)^{n_i} \quad (1)$$

In addition, $\underline{\lambda}$ and $\underline{\phi}$ are independent according to posterior belief as well prior. After some manipulation, (1) becomes

$$\log Posterior(\underline{\phi}) = \quad (2)$$
$$\log Prior(\underline{\phi}) + constant$$
$$- n \cdot HV(\hat{\lambda}_i \hat{\phi}_i, \hat{\lambda}_i(1-\hat{\phi}_i); \lambda_i \phi_i, \lambda_i(1-\phi_i))$$

where $HV$ is a measure of discrimination information [Shore 1985], $\underline{\lambda}$ is any arbitrary proportions and $\hat{\underline{\lambda}}$ and $\hat{\underline{\phi}}$ are the maximum likelihood proportions given the sample statistics $(p_i, n_i)$ for $i=1..C$. That is, $\hat{\lambda}_i = (p_i+n_i)/n$ and $\hat{\phi}_i = p_i/(p_i+n_i)$ for $i=1..C$.

Formula (2) can be interpreted as follows [Cheeseman 1984]. The first term on the right hand side is proportional to the *complexity* of the rule $\underline{\phi}$ in an efficient encoding according to prior belief and the $HV$ term is a measure of the *error* between the rule $\underline{\phi}$ and the observed proportions in the data. So posterior belief in a rule is a trade off between the prior complexity and its observed error weighted by the size of the training set.

### The Best Decision

The induction system is to decide whether each type of object should be classed as positive or negative. Such a decision can be represented by a decision tree with tests on attributes at its nodes and *positive* or *negative* at its leaves. The cost in terms of probability of error of a particular decision given that $(\underline{\lambda},\underline{\phi})$ are the true proportions is[1]

$Cost(decision,(\underline{\lambda},\underline{\phi})) =$
$$\sum_{i=1..C} \lambda_i \left( (1-\phi_i) \cdot 1_{decide\ i\ is\ +ve} + \phi_i \cdot 1_{decide\ i\ is\ -ve} \right)$$

The posterior expectation of this cost represents the expected number of errors. It needs to be minimized. With some manipulation using the posterior independence of $\underline{\lambda}$ and $\underline{\phi}$, it can be shown we should decide to classify an object of type $i$ as a positive instance if and only if $E_{\underline{\phi}}(\phi_i) \geq 1 - E_{\underline{\phi}}(\phi_i)$, that is, $E_{\underline{\phi}}(\phi_i) \geq 0.5$.

### Refinement

Though at this point a statistician would have considered the analysis complete (apart from the choice of prior), further work remains for the purposes of knowledge acquisition. The final result should be presented in a manner comprehensible to the expert. A final *refinement* process is needed to convert proposed decisions to *knowledge*, for example, comprehensible rules. Where comprehensibility is at the cost of a potential decrease in accuracy [Quinlan 1986b], the two processes of decision analysis and knowledge refinement further interact. This interaction is not investigated here.

## 4. CHOOSING A PRIOR

Suppose there is no prior belief correlating the classifications of different types of objects. Then by prior independence and relation (1), posterior belief in the positive classification of an object of type $i$ would be

$$Posterior(\phi_i) \propto Prior(\phi_i) \cdot \phi_i^{p_i}(1-\phi_i)^{n_i}$$

---

[1] The characteristic function $1_A$ equals 1 when $A$ is true and 0 otherwise.

Consequently, only examples of the $i$-th type would reveal further information about $\phi_i$. This means for $C \approx 32,000$, hundreds of thousands of examples are typically needed to reach a reasonable decision if the problem is dealing with uncertainty, and tens of thousands if the problem is in a logical domain (where a priori each $\phi_i$ is either 0 or 1).

These numbers contrast with the training set sizes discussed in Section 2. Yet decision tree induction systems are recording successes with these limited training sets. This is experimental confirmation that in the typical knowledge acquisition problem significant correlation is often justified a priori between different types of objects and decision tree induction systems must be implicitly taking advantage of this. I investigate this second issue in detail in the next section. It is a justification of the first issue that concerns us here.

In the general framework outlined in Section 2, the only problem input that could affect prior belief is the experts' choice of language. The experts believe the attributes chosen are "useful" for classification. If they are basing this on a partial theory or they have shown a credible performance in the past (not just on the training set provided), there is reason to support their belief in the usefulness of the attributes. But, how does this "usefulness" translate into a prior? Clearly, there is no simple or precise answer to this question.

The study of cognitive heuristics in behavioural decision theory has led to some understanding of how experts think [Tversky and Kahneman 1974, Cleaves 1986]. A key heuristic is *representativeness*. In the current context, this says an expert will attempt to classify objects similarly if they are similar in the expert's view. Though in general this heuristic has no statistical basis, if we have some reason to trust the experts' judgement we should place some credibility in the heuristic relative to the language proposed. Common sense tells us that the experts will have taylored their language so that this heuristic is indeed applicable. Afterall, similarity is a language dependent concept [Watanabe 1969, Thm. 7.20].

Another argument goes as follows. Knowing that attributes are useful indicates they have good *discriminating power*. When determining class, this could mean few attributes need to be tested or the cognitive process involved is relatively simple. This suggests that the simpler of two hypotheses should be preferred, all else being equal. Rendell uses this kind of argument to justify selective induction [1986].

The kinds of prior knowledge the above arguments suggest—similar objects are more likely to have similarly classes, and simpler hypotheses are preferred—are not ar odds with each other. They often occur simultaneously. Due to the level of detail that the above arguments support, they may as well be treated as one and the same. I shall refer to either kind of prior, when made on the basis of trust in the expert's judgement, as the *similarity hypothesis*. This is "subjective" Bayesianism taken to the extreme, but when faced with a paucity of data it represents the best use of scant resources. It is used in the following analysis.

## 5. ANALYSING DECISION TREE INDUCTION SYSTEMS

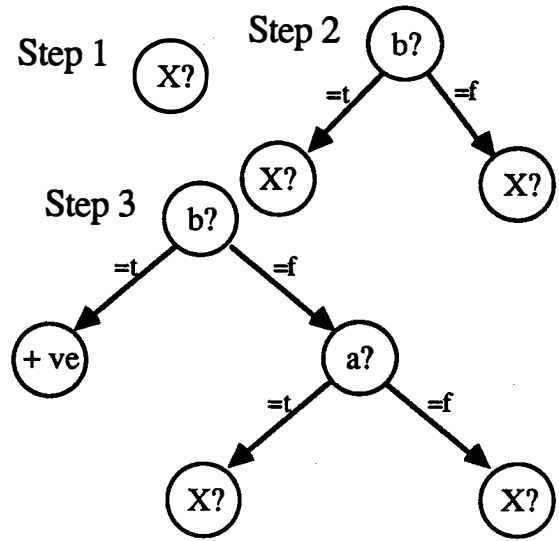

*Figure 1.* Growing a tree.

The basic decision tree approach is as follows. Trees are grown in a top-down manner. Starting from a tree consisting of only one node, each leaf is repeatedly turned into a branch node by placing a test for one of the attributes at the node and constructing branches from it corresponding to each outcome of the test. Fig. 1 demonstrates three initial steps in this process. In step 2, a test on attribute $b$ has been added at the root node and the two leaves below remain to be grown. During the process, a heuristic called a *splitting rule* is used to select an attribute to form the test at the node being grown. Both Breiman *et al.* [1984, p. 102] and Quinlan [1986a, p. 90]

193

suggest as a splitting rule an information-based heuristic that Quinlan calls *gain*[2].

For problems not involving uncertainty, arguments in favour of this basic approach take many forms [Quinlan 1986a, Rendell 1986] but ultimately boil down to the following: simpler trees that fit the data are better. Bayesian analysis says that the approach can only yield repeated success when simpler trees can be expected to perform adequately in the long run on the class of problems being tackled. That is, the similarity hypothesis must hold.

When a decision tree system is applied to problems involving uncertainty, it has been found experimentally that the trees grown using the basic approach can have their accuracy improved by afterwards *pruning* some sections of the tree [Breiman *et al*. 1984, Chap. 3, Quinlan 1986a, 1986b, Quinlan *et al*. 1986]. This is achieved by turning subtrees into leaves and then using some function to assign classes to the new leaf nodes. Fig. 2 demonstrates a simple application. Quinlan also uses pruning for refining knowledge; in this case the aim is to decrease the complexity of the tree without significantly decreasing the accuracy. It is a fortunate coincidence that pruning can sometimes lead to more accurate trees and perform the knowledge refinement process as well. The two aims should be separated out in analysis however. I consider only the former aim here.

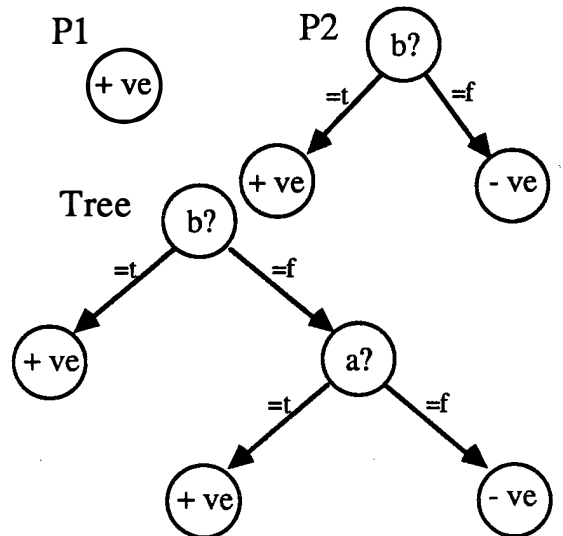

*Figure 2.* Pruning a tree[3].

---

[2] It is also an estimate of the mutual information [Shore 1986] between the class and the attribute.

[3] *P1* is obtained by pruning *Tree* at the node

Using maximum likelihood statistics (by removing the complexity term in equation (2)), pruning cannot be justified as a means of reducing errors. For example, suppose there is a single negative example of a certain type of object and all similar objects in the training set are classified as positive. So when a tree is grown using the standard approach, the negative example would appear as a single negative branch and all neighbouring branches would be positive. Pruning would often discard this single negative branch and absorb it into a neighbouring positive branch. By a maximum likelihood approach, however, only one negative example of that particular type has been seen so the best decision (in terms of minimum errors) must be negative. In fact, by a related argument, pruning as a means of improving accuracy can only be justified under a Bayesian analysis when the prior conforms to something like the similarity hypothesis.

Can we justify decision tree methods from a Bayesian argument? Instead of explaining these methods in detail, I shall develop a similar method from first principles and compare it with the experimentally determined methods.

Rather than evaluating $E_{\underline{\phi}}(\phi_i) \geq 0.5$, consider making a decision whether objects of the $i$-th type should be positive based on whether $\phi_i \geq 0.5$ for some classification rule $\underline{\phi}$ in which there is high posterior belief. This is a reasonable test of whether $E_{\underline{\phi}}(\phi_i) \geq 0.5$ because, first, prior belief should be symmetric about $\phi_i = 0.5$, and second, posterior belief is obtained by weighting the prior by a unimodal likelihood term (the error term in equation (2) raised to the power 2). As the number in the training set increases, this estimate is guaranteed to approach the true value because the error term will dominate.

### The Splitting Rule

For an $N$ attribute problem there are some $2^{2^N}$ possible decision trees. To search this space, the greedy top-down approach advocated by Breiman *et al.* and Quinlan can be adopted. A splitting rule, a heuristic, is needed to guess which attribute should be tested at each step.

Suppose we did test on a particular attribute; consider the tree that results. Fig. 3 represents a tree after a new test on some attribute $X_j$ has been added, resulting in two new leaf nodes

---

testing *b*, and *P2* by pruning at the node testing *a*. The method of reassigning a class to a leaf node has not been shown.

below it. A family of rules (each in the form $\phi$) can be constructed from this tree by assigning different proportions to each leaf-node to representing the chance that an object at the node will be positive (assume the proportions are constant for all object types that would appear at one node). Each of these rules is referred to as an *underlying* rule. Fig. 4 represents such a rule for the tree in Fig. 3.

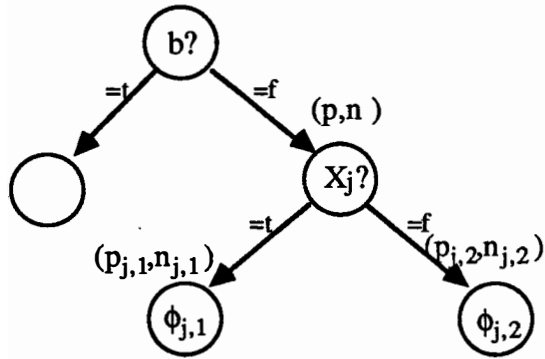

*Figure 3.* Considering the *j*-th attribute.

A 1-ply lookahead heuristic is as follows: select the attribute that leads to a tree with an underlying rule $\phi$ of greatest posterior probability. Under reasonable assumptions concerning the prior, this heuristic is shown below to be the same as selecting the attribute maximizing Quinlan's gain. Notice, however, that the set of underlying rules generated from using, say, a four-valued attribute as the new test will have greater dimensionality (because of four branches rather than two) and consequently more flexibility to match the data by chance than for a binary attribute. Not surprisingly, it has been reported that the gain heuristic is biased towards multi-valued attributes [Kononenko, Bratko and Roskar 1984].

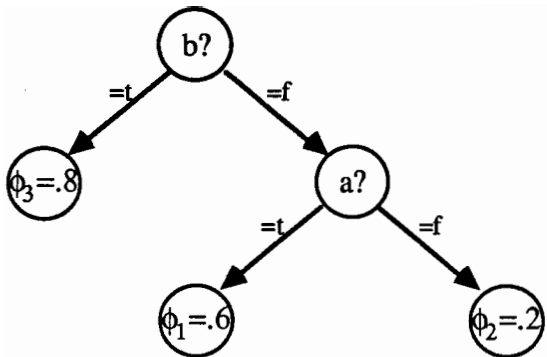

*Figure 4.* An underlying rule.

The relative posterior probability of underlying rules can be determined using (2). If the prior probability of a rule $\phi$ is determined only by the "simplicity"/"adherence to similarity" of the rule or its underlying decision tree, then it will be relatively constant for all rules considered as they have equivalent underlying trees except for a different attribute at one node. So the contribution of the prior can be ignored. Certainly, it can be when the number of examples occurring at the node are large, as the error component of the posterior will dominate.

Only the error terms need be compared. In fact, by the nature of the sum in the error term, only the contribution made by the new test and its leaves need be compared. Suppose the *j*-th attribute is used as the test. Let $(p_{j,1}, n_{j,1})$ be the number of positive and negative examples respectively occurring at the "yes" leaf node of the new test, $(p_{j,2}, n_{j,2})$ the numbers at the "no" node, and $\phi_{j,i}$ the proportions assigned to these nodes by an underlying rule. This situation is represented in Fig. 3. The contribution to the error term then becomes (taking the logarithm of the corresponding term in relation (1), and assuming all attributes are binary)

$(1/n) \sum_{i=1,2} p_{j,i} \cdot \log \phi_{j,i} + n_{j,i} \cdot \log(1-\phi_{j,i})$

The maximum of this w.r.t. $\phi_{j,i}$ for the *j*-th attribute is

$(1/n) \sum_{i=1,2} ( p_{j,i} \cdot \log(p_{j,i}/(p_{j,i}+n_{j,i}))$
$+ n_{j,i} \cdot \log(n_{j,i}/(p_{j,i}+n_{j,i})))$

which is a constant, monotonic function of Quinlan's gain of the *j*-th attribute. So to find the attribute that when used has the tree with an underlying rule $\phi$ of greatest posterior probability, the attribute with the maximum gain should be chosen.

### Stopping and Pruning

When should tree growing be stopped? As a tree is grown, the maximum posterior probability of any underlying rule will not necessarily increase to a maximum and then decrease. The tree may have to reach a certain level of complexity before any underlying rule will show a significant decrease in error. In otherwords, a certain number of questions may need to be asked before any reasonable estimate of the class can be made. The parity problem is an extreme example[4]. Knowing only 7 bits of an 8-bit word gives no clue whatsoever about the parity of the word. So growing cannot be stopped once the maximum posterior probability of any underlying rule for the tree decreases.

---

[4] Incidentally, parity is also an extreme counterexample of "decision tree simplicity" and "adherence to similarity".

But the tree can be grown to its full extent using the gain heuristic and then the best nodes for pruning can be found in hindsight. The resultant tree can be pruned back to the subtree with an underlying rule of maximum posterior probability. This is the hallmark of the various pruning approaches.

Finally, how is the best subtree to be chosen? This presents a problem as a precise form of the prior would need to be specified. Quinlan [1986b] surveys a number of pruning approaches that essentially trade off error with some measure of complexity, as equation (2) suggests. In practice these would best be parameterised and subject to some form of sensitivity analysis, due to the imprecision of the prior.

## 6. DISCUSSION

A number of improvements to decision tree methods follow from the preceeding analysis.

Predicting class proportions at decision tree nodes is a common decision task not yet handled by present ID3-based techniques. Analysis similar to that in Section 3 shows reliable class proportions may need to be estimated when a non-uniform cost structure must be taken advantage of.

An analysis by Breiman *et al.* [1984] shows the difficult nature of the problem. Obstacles to developing such methods are several. The modal estimate given in Section 5 is a rough and ready means of estimating when $E_\phi(\phi_i) \geq 0.5$ but no such estimate for $E_\phi(\phi_i)$ has yet been proposed. Applying Laplace's law of succession at the nodes is not appropriate because of the way the trees have been grown to fit the data [Buda, personal communication, Niblett and Bratko 1987]. The usual Bayesian method of evaluation is to approximate a prior using the natural conjugate to the updating formula given in (1) [Raiffa and Schlaifer 1961, Howard 1970]. This is not possible under the similarity hypothesis as the natural conjugate is unimodal, whereas the prior favouring similarity cannot be, according to any reasonable interpretation of similarity or simplicity.

A potential solution is to obtain an estimate of $E_\phi(\phi_i)$ by stochastically generating several rules of high posterior belief and pooling their $\phi_i$. This approximates $E_\phi(\phi_i)$ because it estimates the contribution from the most dominant rules in the expectation.

A revolutionary method of improving the decision tree approach is too adopt a better model of simplicity. Simplicity of sets of conjunctive rules is, in most cases, a more natural measure of "simplicity" than simplicity of trees. For instance, a disjunction of a few simple conjunctions can translate into quite a complex tree. Quinlan suggests [1986b] developing rule sets instead of trees by a method based on the known, efficient tree-building technology. Alternatively, a bottom-up method based on common AI generalization techniques could be used. This has the added advantage that it sidesteps the problem of splitting rules altogether (although at added computational cost) and allows background knowledge to be more flexibly incorporated.

## 7. CONCLUSION

Simplicity and similarity are key issues in induction. Bayesian updating is language independent so these issues can only become relevant either through the choice of prior or through some argument involving computational constraints. The computational constraints exist both because of cognitive constraints on the induction product itself, mentioned in Section 2, and because of restrictions imposed on the induction procedure or the class of hypotheses being considered [Pearl 1978, Rendell 1986]. We have focussed on the choice of prior. We have considered the justification and effect of a prior favouring simplicity.

The Bayesian interpretation of decision tree methods, utilising this prior, serves to explain phenomena such as the practice of pruning, the bias by Quinlan's gain heuristic towards multi-valued attributes and the problems with estimating proportions at nodes.

Furthermore, the similarity hypothesis (proposed in Section 4 and shown in Section 5 to be implicity assumed for decision tree pruning) clearly delimits the kinds of applications to which decision tree methods should be applicable.

## ACKNOWLEDGEMENTS

These ideas have been developed through numerous discussions with Renato Buda, Chris Carter and other members of the AI laboratory at NSWIT. Ross Quinlan's many research initiatives provided the incentive. Thanks are also due to Peter Cheeseman for indicating references and the workshop program committee for changes suggested.

## REFERENCES

Breiman, L., Friedman, J.H., Olshen, R.A. and Stone, C.J (1984). *Classification and Regression Trees.* Belmont: Wadsworth.

Cheeseman, P.C. (1984). Learning expert systems from data. *IEEE Workshop on Principles of knowledge-based systems*. Denver. pp. 115-121.

Cleaves, D.A. (1986). Cognitive biases and corrective techniques: proposals for improving elicitation procedures for knowledge-based systems. *Knowledge Acquisition for Knowledge-Based Systems Workshop*, Banff, Canada. To appear in *International Journal for Man-Machine Studies*.

Horvitz, E.J., Heckerman, D.E. and Langlotz, C.P. (1986). A framework for comparing alternative formalisms for plausible reasoning. *Fifth National Conference on Artificial Intelligence*. Philadelphia. pp. 210-214.

Howard, R.A. (1970). Decision analysis: perspectives on inference, decision, and experimentation. *Proceedings of the IEEE* **58**(5).

Kononenko, I., Bratko, I. and Roskar, E. (1984). Experiments in automatic learning of medical diagnostic rules (Technical Report). Josef Stefan Institute, Ljubljana, Yugoslavia.

Michalski, R.S. (1983). A theory and methodology of inductive learning. *Artificial Intelligence* **20**. pp. 111-161.

Michie, D. (1986a). The superarticulacy phenomenon in the context of software manufacture. To appear in *Proc. Roy. Soc. (A)*.

Michie, D. (1986b). Current developments in expert systems. In: Quinlan, J.R. (Ed.), *Applications of Expert Systems*. London: Addison Wesley.

Niblett, T. and Bratko, I. (1987). Learning decision rules in noisy domains (Technical Report). Turing Institute, Glasgow, Scotland.

Paterson, A. and Niblett, T. (1983). *ACLS user manual*. Glasgow: Intelligent Terminals Ltd.

Pearl, J. (1978). On the connection between the complexity and credibility of inferred models. *International Journal of General Systems*, **4**.

Quinlan, J.R. (1986a). Induction of decision trees. *Machine Learning* **1**(1).

Quinlan, J.R. (1986b). Simplifying Decision Trees. *Knowledge Acquisition for Knowledge-Based Systems Workshop*, Banff, Canada. To appear in *International Journal for Man-Machine Studies*.

Quinlan, J.R., Compton, P.J., Horn K.A. and Lazarus, L. (1986). Inductive knowledge acquisition: a case study. *Proceedings of the Second Australian Conference on Applications of Expert Systems*. pp. 183-204.

Raiffa, H. and Schlaifer, R. (1961). *Applied Statistical Decision Theory*. Harvard: Harvard University.

Rendell, L. (1986). A general framework for induction and a study of selective induction. *Machine Learning* **1**(2). pp. 172-226.

Shore, J.E. (1985). Relative entropy, probabilistic inference and AI. *Uncertainty in AI Workshop*. UCLA.

Tversky, A. and Kahneman, D. (1974). Judgements under uncertainty: heuristics and biases. *Science* **185**. pp. 1124-1131.

Watanabe, S. (1969). *Knowing and Guessing: A Formal and Quantitative Study*. New York: John Wiley.